\documentclass[letterpaper, 10 pt, conference]{ieeeconf}  

\IEEEoverridecommandlockouts                              

\title{\LARGE \bf
\textit{NeuroEdge}: Real-Time Hand Gesture Recognition with High-Density EMG Using Deep Learning  at the Edge
}
\author{Peter~Chudinov$^{1}$, Zhenyu~Lin$^{2}$, Jay~Motamarry$^{3}$, Srihita~Panati$^{4}$, Xiaorong~Zhang$^{2}$, and Zhuwei~Qin$^{2}$
\thanks{This work was supported by Sony Sensing Solution University Program.}
\thanks{Department of Biology$^{1}$ and School of Engineering in Computer Engineering$^{2}$ at San Francisco State University, San Francisco, CA 94132.
College of San Mateo$^{3}$, CA 94402. Contra Costa College$^{4}$, CA 94806. Corresponding authors e-mail: {peterchudinov@icloud.com, zwqin@sfsu.edu}}
}

\usepackage{cite}
\usepackage{amsmath,amssymb,amsfonts}
\usepackage{algorithmic}
\usepackage{graphicx}
\usepackage{textcomp}
\usepackage{xcolor}
\def\BibTeX{{\rm B\kern-.05em{\sc i\kern-.025em b}\kern-.08em
    T\kern-.1667em\lower.7ex\hbox{E}\kern-.125emX}}
    
\begin{document}

\maketitle
\thispagestyle{empty}
\pagestyle{empty}


\begin{abstract}
High-density electromyography (HD-EMG) has emerged as a powerful modality for decoding fine-grained neuromuscular activity, enabling real-time neural-machine interfaces (NMIs) for applications such as prosthetic control, rehabilitation, and augmented interaction. 
While deep learning approaches such as convolutional neural networks (CNNs) have demonstrated high classification accuracy for EMG-based gesture recognition, their deployment on embedded hardware remains a major challenge due to computational and memory constraints. 
This paper presents \textit{NeuroEdge}, a real-time HD EMG-based NMI system that performs gesture recognition entirely on resource-constrained microcontrollers. 
The system features two custom-designed modules: the HD-EMG StreamBridge, a wireless communication interface that streams raw HD-EMG data from a Quattrocento amplifier to an ESP32 microcontroller; and the EdgeDL Inference Engine, a lightweight deep learning framework executing on a Sony Spresense microcontroller. 
A compact 1-dimensional CNN optimized for embedded inference processes, sliding windows of EMG data in real time. 
Data streaming and inference are pipelined and synchronized through an architecture that utilizes Direct Memory Access (DMA) for data transfer and Serial Peripheral Interface (SPI) burst communication between the ESP32 and Spresense, ensuring low-latency performance. 
Experimental results show that \textit{NeuroEdge} achieves a real-time classification accuracy of 90\% across seven hand gestures, with a total average latency of 83 ms using 192 channels of HD-EMG recorded from the forearm. 
Our system demonstrates the feasibility of deploying complex HD-EMG-based gesture recognition on microcontroller-based edge devices, bridging the gap between high-resolution biosignal acquisition and deep learning-based embedded inference for next-generation NMIs. 
\end{abstract}


\section{Introduction}
Neural-machine interfaces (NMIs) based on surface electromyography (EMG) signals offer a non-invasive approach for decoding motor intent, with promising applications in prosthetic control, rehabilitation robotics, exoskeletons, and virtual interaction systems~\cite{Zhang, Zirbel}. 
Traditional EMG-based NMIs employ sparse electrode configurations with feature engineering that transform raw EMG data into meaningful and discriminative features that can be used for tasks like classification and regression. 
While these approaches are computationally efficient, hand-crafted features may not generalize well across different tasks, subjects, or recording conditions.

Recent advances in high-density EMG (HD-EMG) and deep learning have markedly improved gesture recognition performance. 
HD-EMG systems use multi-electrode arrays to capture detailed neuromuscular activity, while deep neural networks, including convolutional (CNNs), recurrent (RNNs), and hybrid models, leverage these rich spatiotemporal features for accurate motor intent decoding~\cite{Chamberland}. 
However, deploying such systems in real-world, portable scenarios remains challenging due to their computational demands.

Several prior studies have attempted to bridge this gap. 
Tam et al. developed a fully embedded HD-EMG gesture classification system, which includes a custom-built 32-channel flexible electrode array and a compact CNN model deployed on a GPU-equipped NVIDIA Jetson Nano~\cite{Tam}. 
Buteau et al. used the Coral TPU and 64-channel HD-EMG sensors for TinyML-based gesture recognition, enabling real-time inference with on-device fine-tuning~\cite{Buteau}. 
Dere et al. deployed an event-driven neural network on an FPGA using hybrid EMG and EEG inputs (8 channels each) for gesture classification~\cite{OA}. 
While these systems demonstrated strong promise and performance, their reliance on power-hungry platforms, such as GPUs, TPUs, and FPGAs, limits their portability and suitability for real-time applications that require low-power, low-cost microcontroller-based solutions.

Prior work has explored network compression techniques, including model quantization and pruning, to enable inference on resource-constrained microcontroller devices. 
Lu et al. demonstrated the feasibility of deploying a CNN-based EMG recognition model on the Sony Spresense microcontroller, utilizing 8-channel EMG data along with 8-bit quantization and transfer learning to enable efficient inference on resource-constrained hardware~\cite{Lu}. 
Just et al. also investigated neural network quantization on microcontrollers, offering valuable insights into efficient deployment under memory and compute constraints~\cite{Just}. 
However, these efforts focused primarily on model optimization and did not address real-time HD-EMG data acquisition or end-to-end embedded system integration. 

In this work, we introduce \textit{NeuroEdge}, a real-time hand gesture recognition system that integrates HD-EMG, wireless data transmission, and embedded deep learning inference on resource-constrained microcontrollers. 
At the core of the system are two key modules: (1) the HD-EMG StreamBridge, a custom-designed wireless communication interface that streams raw HD-EMG data from a Quattrocento EMG system with up to 384 channels to an ESP32 microcontroller; and (2) the EdgeDL Inference Engine, a lightweight deep learning framework running on a Sony Spresense microcontroller. 
A compact 1D CNN, adapted from the M5 voice recognition model and optimized for embedded inference, processes sliding windows of EMG data in real time~\cite{Dai}. 
Data streaming and inference are pipelined and synchronized through an architecture that utilizes Direct Memory Access (DMA) for data transfer and Serial Peripheral Interface (SPI) burst communication between the ESP32 and Spresense, ensuring low-latency performance. 
To the best of our knowledge, \textit{NeuroEdge} is the first demonstration of full-scale HD-EMG deep learning inference on microcontroller hardware with real-time performance capabilities.  

\section{Methods}
\subsection{\textit{System Overview}}

\begin{figure}[t!]
	\begin{center}
	\vspace{2mm}
	\includegraphics[width=3.5in]{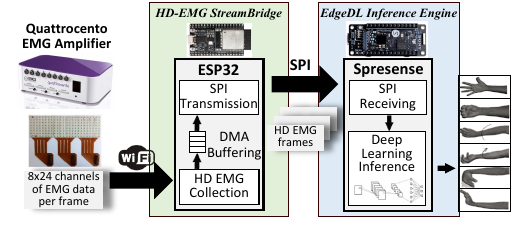}
	\vspace{-6mm}
	\caption{Overall Architecture of the \textit{NeuroEdge} System.}
	\label{fig1}
	\end{center}
	\vspace{-6mm}
\end{figure}

Fig.~\ref{fig1} illustrates the overall architecture of the \textit{NeuroEdge} system, which comprises two main functional modules: 1) the \textbf{\textit{HD-EMG StreamBridge}}, a custom-designed wireless interface enabling real-time, high-throughput communication between the HD-EMG amplifier Quattrocento (OT Bioelettronica, Italy) and the ESP32 microcontroller, which features built-in Wi-Fi; and 2) the \textbf{\textit{EdgeDL Inference Engine}}, a lightweight deep learning module that performs real-time gesture recognition on a Sony Spresense microcontroller using a compact 1D CNN model. 
HD-EMG frames are transmitted from the HD-EMG StreamBridge (based on the ESP32) to the EdgeDL Inference Engine (based on the Spresense) via high-speed SPI communication. 
All communication and computing components are seamlessly integrated to support real-time gesture recognition on edge devices.

While the \textit{NeuroEdge} system is not restricted to a fixed number of EMG input channels, in our demonstration prototype, 192 channels of EMG signals were collected from the Quattrocento amplifier at a sampling rate of 512 Hz for hand gesture recognition. These signals were processed by the EdgeDL Inference Engine in sliding windows of 20 frames. These parameters will be referenced in the system design discussion in the following sections. Details of the data collection protocol are provided in Section III.A.

\subsection{\textit{HD-EMG StreamBridge}}

While the Quattrocento EMG system includes MATLAB-based software for raw EMG data acquisition, it lacks compatibility with microcontrollers. Furthermore, it requires an Ethernet cable connection to communicate with a PC. 
To overcome these limitations, we developed the HD-EMG StreamBridge, a custom Wi-Fi-based wireless communication interface designed specifically for the Quattrocento HD-EMG system.
The StreamBridge includes two versions:
\begin{itemize}
\item \textbf{HD-EMG StreamBridge-Python:}  
A Python-based interface for communication with PCs and embedded platforms with operating systems (e.g., Raspberry Pi).
\item \textbf{HD-EMG StreamBridge-C:}
A C-based interface for communication with ESP32 microcontrollers.
\end{itemize}

These interfaces provide a robust and scalable solution for integrating HD-EMG data into portable and wearable devices. 
Communication with the Quattrocento system follows the Transmission Control Protocol (TCP) and is initiated via a 40-byte command structure. 
The process begins with the PC or ESP32 sending a 40-byte command string that includes parameters such as sampling frequency, channel configuration, filter settings, analog output selections, and detection modes. This command also includes a CRC-8/MAXIM checksum to ensure data integrity.
Once the Wi-Fi communication is established and the command is transmitted, Quattrocento begins sampling EMG signals from the specified channels at the configured sampling frequency. The device buffers the acquired data and streams it to the connected host.

For \textbf{HD-EMG StreamBridge-Python}, we developed an open-source Python library for efficient command byte generation, CRC-8/MAXIM checksum computation, and real-time data streaming. The library is available at: https://github.com/screamuch/quattrocento-python (DOI: 10.5281/zenodo.7869982).

For \textbf{HD-EMG StreamBridge-C} running on the ESP32 microcontroller, the ESP32 operates as a Wi-Fi client while the Quattrocento functions as the server. The ESP32 initiates serial communication at a baud rate of 115200. Once a wireless TCP/IP connection is established with the Quattrocento, the ESP32 sends specific 40-byte command sequences (as previously described) to initiate EMG data collection.

To ensure Quattrocento starts in a stable state, ESP32 issues a stop command to clear any residual activity from previous sessions, followed by a start command to begin a fresh acquisition cycle. 
It discards the first 1,000 samples to minimize synchronization issues and then streams the 192 16-bit active channels.
A data-shifting strategy is implemented within the buffer to maintain the temporal order of samples, supporting continuous, real-time signal processing.


In the experimental prototype, HD-EMG data is processed in sliding analysis windows for gesture recognition. Each window comprises 20 samples across 192 channels, forming a 7,680-byte data block (20 × 192 × 2 bytes/sample). This window is buffered in the ESP32’s on-chip memory using DMA, allowing efficient, nonblocking data storage. Once a full window is buffered, the ESP32 initiates communication with Spresense through a simple handshake protocol. In this configuration, the Spresense serves as the SPI controller and the ESP32 as the peripheral. Upon receiving a ready signal from the ESP32, the Spresense retrieves the buffered data through SPI burst transmission.

\begin{figure}[t!]
	\begin{center}
	\vspace{2mm}
	\includegraphics[width=3.5in]{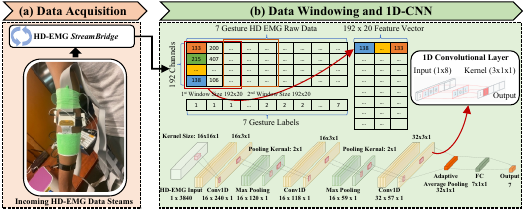}
	\vspace{-6mm}
	\caption{HD-EMG Data Windowing and 1D CNN-based Inference.}
	\label{fig2}
	\end{center}
	\vspace{-6mm}
\end{figure}

\subsection{\textit{EdgeDL Inference Engine}}

The EdgeDL Inference Engine was implemented on a Sony Spresense to perform real-time gesture classification from HD-EMG data using a lightweight 1D CNN Model optimized for embedded deployment. The Spresense is a multicore ARM Cortex-M4F–based microcontroller optimized for low-power, real-time AI workloads. It is well-suited for embedded deep learning due to its 1.5 MB of on-chip RAM and support for TensorFlow Lite for microcontrollers.


As shown in Fig.~\ref{fig2} (b), the HD-EMG data are processed in sliding windows for gesture recognition. Each window consists of 20 samples across 192 channels. Upon receiving a window of data from the StreamBridge module via SPI, the Spresense converts incoming int16 EMG values to float32 format to meet the input requirements of TensorFlow Lite.

The inference model is a 1D CNN adapted from the M5 architecture, originally developed for audio classification. 
The model input is a 1×3840 feature vector, formed by flattening a 20×192 matrix representing 20 consecutive EMG time samples from 192 channels. This structure preserves both temporal and spatial characteristics of muscle activation.

The CNN architecture consists of:

\begin{itemize}
\item A first convolutional layer with 16 filters (kernel size: 16, stride: 16), reducing temporal resolution and extracting spatial features aligned with the 8 × 8 sensor grid;
\item A second layer with 16 filters (kernel size: 3, stride: 1);
\item A third layer with 32 filters, each followed by max pooling (stride: 2) to downsample and introduce translational invariance.
\end{itemize}

To minimize memory footprint, the model uses a fully convolutional design, omitting dense layers. A global average pooling layer precedes a final fully connected layer, yielding a compact representation suitable for classification within the constraints of embedded deployment. A dedicated tensor arena is allocated to manage model computations and memory reuse efficiently.

During inference, the Spresense invokes the 1D CNN model to perform inference, logging inference durations for performance profiling. The output probabilities are parsed to identify the gesture class with the highest likelihood, which is recorded for debugging and downstream analysis.

Model training was conducted in PyTorch, using Conv2d layers with degenerate spatial dimensions to efficiently process sequential EMG data. For deployment on the Spresense, the trained PyTorch model was exported to ONNX, converted to TensorFlow Lite, and compiled into a C header file for integration with the microcontroller firmware. This end-to-end deployment pipeline supports real-time, on-device gesture inference—from model training to embedded execution.

To support continuous, low-latency processing, the system employs a pipelined architecture. Data acquisition on the ESP32, data transfer via SPI, and gesture inference on the Spresense are executed concurrently. This design ensures seamless integration of signal collection and deep learning inference, minimizing latency and maximizing throughput in real-time EMG processing.

\section{Experiments}
\subsection{\textit{Data Collection Protocol}}

This study was conducted with Institutional Review Board (IRB) approval at San Francisco State University (SFSU) and with the informed consent of the subject. Data collection involved one able-bodied male subject using three 8×8 electrode arrays (GR10MM808, OT Bioelettronica, Italy), with electrodes spaced evenly at 10 mm intervals, for a total of 192 channels. These arrays were positioned on the subject’s dominant forearm near the elbow to target muscles responsible for hand, palm, and finger movements (Fig.~\ref{fig2} (a)). Foam pads were saturated with AC Cream before electrode placement. Electrodes were secured to the skin using adhesive bandages, with reference straps placed at the elbow (ground) and approximately 2.5 cm distal to the electrode array (muscle reference). The reference straps were kept damp throughout the data acquisition period.

Seven hand and wrist gestures were included in the experiment: no movement, wrist supination, wrist pronation, hand close, hand open, wrist flexion, and wrist extension. Data were sampled at 512 Hz, with a 0.3 Hz high-pass and 500 Hz low-pass filter applied.

\subsection{\textit{Experiment Setup}}

The experiment consisted of two phases: offline training and real-time testing.
\textbf{Offline Training:} A Python-based GUI, built on HD-EMG StreamBridge-Python, cued the subject to perform each gesture for 7 seconds, repeated 8 times with rest intervals. HD-EMG data were acquired using the Quattrocento system and labeled in real time on an M2 MacBook Air. The resulting CSV files (192 channels × time) were used to train the gesture classifier. Model training was performed on an NVIDIA RTX 3080 GPU.
\textbf{Real-time Testing:} Immediately following training (to preserve electrode placement), the same gestures were performed in response to visual and verbal cues. During this phase, data were streamed via HD-EMG StreamBridge-C on the ESP32 and processed by the EdgeDL Inference Engine on the Sony Spresense. Classification outputs were sent to a PC via serial communication and displayed in real time. All sessions were video recorded for offline alignment and analysis.


\subsection{\textit{NeuroEdge System Performance}}

\textbf{Offline Performance:} 
The gesture recognition achieved a validation classification accuracy of 95.33\% across seven gesture classes after 25 training epochs in 15 minutes. 
The resulting model is a compact 8-bit integer C header file with a final size of approximately 15.24 KB. 


\begin{figure}[t!]
	\begin{center}
	\includegraphics[width=3.5in]{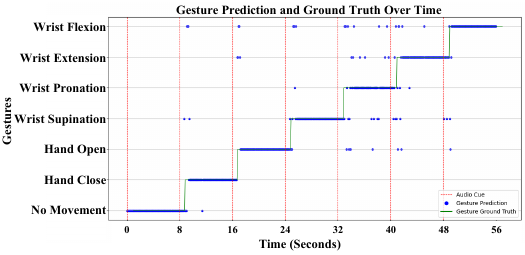}
	\vspace{-6mm}
	\caption{Temporal Alignment of Real-time Gesture Predictions.}
	\label{fig3}
	\end{center}
	\vspace{-6mm}
\end{figure}

\textbf{Real-time Performance:} 
Fig.~\ref{fig3} illustrates the temporal alignment of a representative gesture prediction trial over a 56-second interval. The horizontal axis denotes time, while the vertical axis indicates the seven gesture classes. Red dashed lines mark the timing of synchronized audio-visual cues delivered to the subject at 8-second intervals. Ground truth labels (green lines) are offset by 1.365 seconds to account for human response delay and gesture transition latency, based on synchronized video analysis. 
Model predictions (blue dots) reflect raw classifier outputs without post-processing. Each gesture appears as a horizontal segment, with predicted and true labels overlaid to visualize alignment. The overall classification accuracy across both steady-state and transition phases was 90.00\%.

Most misclassifications occurred during gesture transitions, which is consistent with the fact that training data included only static gesture samples. Incorporating transition data into the training set could potentially improve performance during these phases. 
While post-processing techniques such as majority voting could further enhance classification accuracy, they may introduce additional latency that could compromise real-time responsiveness.
Finally, real-time testing demonstrated an average inference latency of 70 milliseconds and an SPI communication latency of 13 milliseconds.

In summary, the \textit{NeuroEdge} demonstrated the feasibility of real-time gesture recognition using HD-EMG signals and deep learning on a resource-constrained microcontroller. These findings support the potential for deploying edge-based machine learning in wearable and biomedical applications.

\section{Conclusion and Acknowledgments}
This work presents \textit{NeuroEdge}, a fully integrated, real-time neural interface system that combines HD-EMG signal acquisition with embedded deep learning inference on resource-constrained microcontrollers. By introducing the HD-EMG StreamBridge for wireless, high-throughput data streaming and the EdgeDL Inference Engine for on-device classification using a compact CNN, \textit{NeuroEdge} demonstrates the feasibility of deploying complex gesture recognition pipelines directly on low-power hardware. 
To the best of our knowledge, \textit{NeuroEdge} is the first system to enable full-scale HD-EMG inference using deep learning models on microcontroller-based platforms, without reliance on external computing resources. These results underscore the potential of edge-based neural-machine interfaces for next-generation wearable, prosthetic, and human-computer interaction systems.

Future work will explore multi-subject validation, inclusion of gesture transitions in training data, and adaptive learning techniques to enhance robustness across users and sessions. Additionally, expanding the system to support continuous gesture decoding and multi-modal sensor fusion will further increase its applicability in real-world environments.

The authors would like to thank Thandapani Srinivasan Chandrasekaran for his support in the experiments.

\vspace{12pt}

\end{document}